\documentclass[conference]{IEEEtran}
\IEEEoverridecommandlockouts

\usepackage{cite}
\usepackage{hyperref}
\usepackage{amsmath,amssymb,amsfonts}
\usepackage{algorithmic}
\usepackage{graphicx}
\usepackage{textcomp}
\usepackage{multirow}
\usepackage{xcolor}
\usepackage{booktabs}
\def\BibTeX{{\rm B\kern-.05em{\sc i\kern-.025em b}\kern-.08em
    T\kern-.1667em\lower.7ex\hbox{E}\kern-.125emX}}
\begin{document}

\title{NCL-CIR: Noise-aware Contrastive Learning for Composed Image Retrieval}

\author{\IEEEauthorblockN{Peng Gao\IEEEauthorrefmark{1}\IEEEauthorrefmark{2}, Yujian Lee\IEEEauthorrefmark{1}, Zailong Chen\IEEEauthorrefmark{3}, Hui Zhang\IEEEauthorrefmark{2}\IEEEauthorrefmark{5}, Xubo Liu\IEEEauthorrefmark{4}, Yiyang HU\IEEEauthorrefmark{2}, Guquan Jing\IEEEauthorrefmark{1}}
\IEEEauthorblockA{\IEEEauthorrefmark{1}Hong Kong Baptist University, Hong Kong SAR \IEEEauthorrefmark{2} BNU-HKBU United International College, China\\
\IEEEauthorrefmark{3}University of Wollongong, Australia\\ 
\IEEEauthorrefmark{4} University of Surrey, United Kingdom}
\thanks{\IEEEauthorrefmark{5}Corresponding author. This work is supported by the National Natural Science Foundation of China (62076029), National Key R\&D Program of China (2022YFE0201400), Guangdong Science and Technology Department (2022B1212010006), Guangdong Key Lab of AI and Multi-modal Data Processing (2020KSYS007). \textcopyright\ 2025 IEEE. Personal use of this material is permitted. Permission from IEEE must be obtained for all other uses, in any current or future media, including reprinting/republishing this material for advertising or promotional purposes, creating new collective works, for resale or redistribution to servers or lists, or reuse of any copyrighted component of this work in other works. DOI:10.1109/ICASSP49660.2025.10888719}
}


\maketitle

\begin{abstract}
Composed Image Retrieval (CIR) seeks to find a target image using a multi-modal query, which combines an image with modification text to pinpoint the target. While recent CIR methods have shown promise, they mainly focus on exploring relationships between the query pairs (image and text) through data augmentation or model design. These methods often assume perfect alignment between queries and target images, an idealized scenario rarely encountered in practice. In reality, pairs are often partially or completely mismatched due to issues like inaccurate modification texts, low-quality target images, and annotation errors. Ignoring these mismatches leads to numerous False Positive Pair (FFPs) denoted as noise pairs in the dataset, causing the model to overfit and ultimately reducing its performance. To address this problem, we propose the Noise-aware Contrastive Learning for CIR (NCL-CIR), comprising two key components: the Weight Compensation Block (WCB) and the Noise-pair Filter Block (NFB). The WCB coupled with diverse weight maps can ensure more stable token representations of multi-modal queries and target images. Meanwhile, the NFB, in conjunction with the Gaussian Mixture Model (GMM) predicts noise pairs by evaluating loss distributions, and generates soft labels correspondingly, allowing for the design of the soft-label based Noise Contrastive Estimation (NCE) loss function. Consequently, the overall architecture helps to mitigate the influence of mismatched and partially matched samples, with experimental results demonstrating that NCL-CIR achieves exceptional performance on the benchmark datasets.




\end{abstract}


\section{Introduction}
Composed image retrieval (CIR) presents a complex challenge, encompassing the process of searching for a target image using both a reference image and accompanying text that outlines desired modifications\cite{zhen2019deep}. Essentially, the objective of CIR is to locate an image in the gallery that has incorporated the changes indicated by the textual description, all the while maintaining a visual resemblance to the original reference image. Vo et al. first \cite{vo2019composing} propose the TIGR model for deal with CIR. The challenge in composed image retrieval is ensuring semantic consistency and accurate feature alignment between the multi-modal query (reference image coupled with the modification text) and target images. To effectively tackle these challenges, subsequent methodologies in CIR can be delineated into three distinct stages: cross-modal fusion, multi-scale cross-modal fusion, and Vision-Language Pre-trained Models (VLP). In the initial stage, networks \cite{sun2024image,10350916} concentrate on highly effective multi-modal fusion techniques aimed at establishing a robust common latent space for both text and images. Yet, this approach has proven insufficient for effectively tackling CIR issues, as images inherently contain far richer semantic information than text. Consequently, to understand and process the data more comprehensively, the second-stage models such as \cite{pang2022heterogeneous}, further delve deeper into the homogenized semantic information shared between text and images across various scales, Li et al. \cite{10.1145/3639469} propose the cross-modal attention preservation method to solve the issue of correspondence between text-image relationships, and \cite{yang2023composed} propose the cross-modal augmented space for multi-scale matching. However, with the emergence of large models, an increasing number of people are applying contrastive learning - VLP \cite{liu2021image} to CIR. By leveraging large-scale pre-training and unified representation learning, VLP models provide a robust baseline accuracy and facilitate the projection of images and text into a shared feature space for effective computation.





\begin{figure}[!t]
  \makebox[0.5\textwidth]{\includegraphics[scale=0.54]{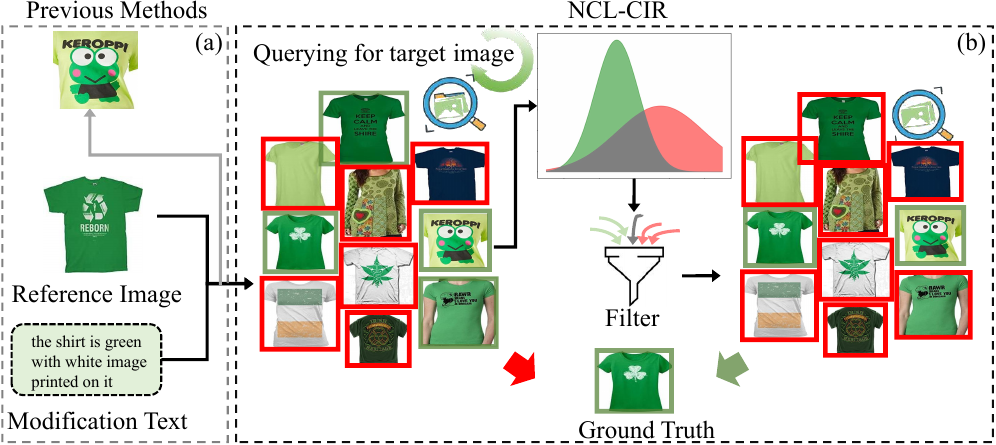}}
  \caption{(a). It presents the problem of partially matched or mismatched pairs existing in the previous methods. (b) A rough workflow of filtering the noise pairs and preserving the matched pairs in the proposed NFB in NCL-CIR.}
  \label{workflow1}
\end{figure}

Although the aforementioned methods achieve superior performance, they all assume all multi-modal queries and target images are perfectly aligned during training, which is clearly an idealized scenario. As shown in Fig. \ref{workflow1}(a), even though the reference image and modification text have detailed corresponding descriptions,  pairing discrepancy can still occur due to the abundance of similar target samples. Different from annotations errors, this term refers to incorrect associations between multi-modal query and target image pairs, where False Positive Paris (FFPs) denoted as noise pairs are involved in training. Essentially, this involves treating negative pairs as if they are positive during the learning process. Including such samples in model training can lead to overfitting to mismatched or partially matched examples and eventually degrade model performance. To address these problems, we proposed the Noise-aware Contrastive Learning for CIR (NCL-CIR) in Fig. \ref{workflow1}(b), composing four components: the CLIP encoder, the Weight Compensation Block (WCB), the Noise-pair Filter Block (NFB), and the soft-label based Noise Contrastive Estimation (NCE) loss function. The main contributions of this paper are three-fold:

1. We propose the NCL-CIR, which takes the matched pairs (multi-modal queries and target images) and the noise pairs (partially matched pairs and mismatched pairs) into consideration, demonstrating remarkable performance.

2. Instead of relying solely on global embeddings from the VLP encoder, the WCB dynamically re-weights diverse embeddings by weight maps, preserving more semantic information from the encoders' outputs.

3. The NFB offers two significant advantages. Firstly, it filters noise pairs (partially matched and mismatched pairs) while preserving the matched pairs. Secondly, NFB generates soft labels for the matched pairs to be utilized in the loss function, resulting in a cohesive approach.

\section{Related Work}
\subsection{Composed Image retrieval}
CIR has garnered growing research attention due to its profound theoretical importance and attractive commercial opportunities. The CIR task is first proposed by Vo et al.\cite{vo2019composing}, which uses the "Text Image Residual Gating" to fuse reference image and modification text. The methods for addressing the CIR problem can be broadly categorized into three types: multi-modal fusion methods, attention-based methods, and contrastive learning methods. Researchers like \cite{anwaar2021compositional, dodds2020modality,liu2021image} try to find the common latent space for reference images and modification texts by late-fusion, which have attempted to bridge the significant gap between the reference image and the modification text, introducing a series of solutions for CIR. Moreover, \cite{chen2020image, delmas2022artemis, liu2024bi, chen2024spirit} make significant contributions by introducing lightweight attention mechanisms that effectively mediate between image and text in multi-modal queries, focusing on their specific relationships with the target image. With the development of pre-trained models, most current CIR methods\cite{goenka2022fashionvlp} now utilize VLP (Vision Language Pre-train) models\cite{radford2021learning, han2022fashionvil, mirchandani2022fad} pre-trained under contrastive learning to handle CIR tasks. These models are highly effective at extracting representations of images and text for downstream tasks, significantly enhancing model performance.
\subsection{Noise-aware Learning}
The noise-aware paradigm is first introduced by \cite{huang2021learning} for cross-modal matching problems, aiming to mitigate the negative effects of mismatched pairs. This specialized learning approach has been widely applied to various cross-modal tasks, including image-text retrieval \cite{zhang2024negative, han2024learning}, visual-audio learning \cite{han2024noise}, and dense retrieval \cite{zhang2023noisy}. Despite the success of previous methods in cross-modal matching, the issues of partial sample matching and noise-aware learning in CIR tasks remain unexplored. In this paper, we present a robust framework for CIR to tackle these noise-aware learning challenges, achieving promising results across three public datasets.

\begin{figure}[!t]
  \makebox[0.5\textwidth]{\includegraphics[scale=0.5]{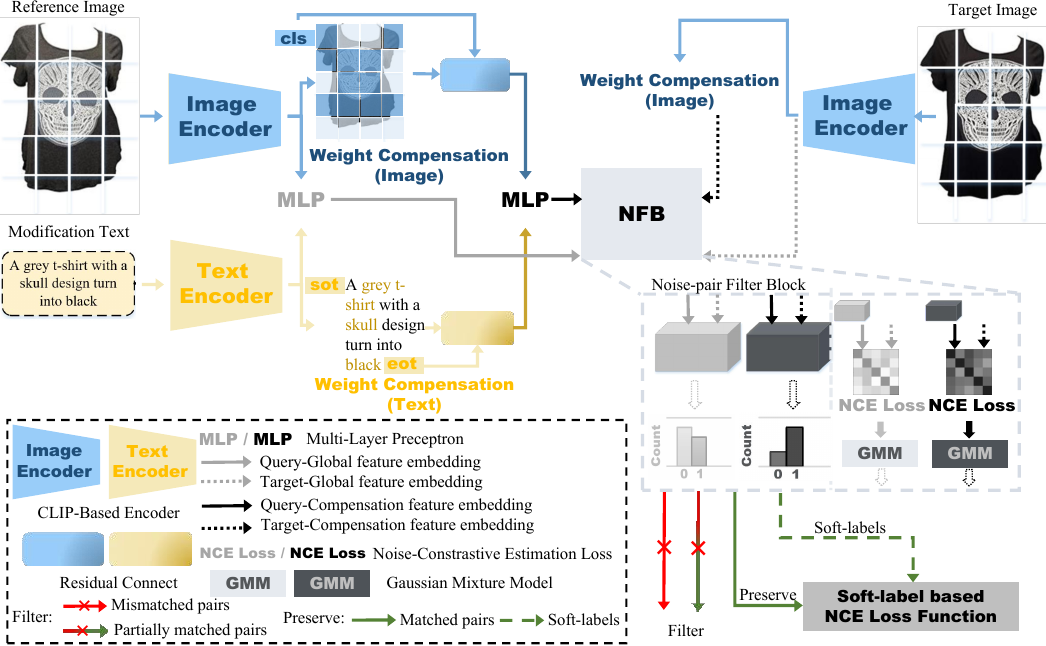}}
  \caption{The workflow of NCL-CIR begins by encoding the modification text, reference image, and target image with CLIP to extract overall features and attention maps. Before the pairs are processed by the Noise-pair Filter Block (NFB), the feature embeddings pass through the Weight Compensation Block (WCB). This step produces refined embeddings, yielding multi-scale pair feature representations that enhance NFB's ability to filter noise pairs and preserve matched pairs. Additionally, NFB generates soft labels for the matched pairs, facilitating improved training within the soft-label based Noise Contrastive Estimation (NCE) loss function.}
  \label{workflow2}
\end{figure}
\section{Methodology}
\label{second}
This section first formulates the token embeddings obtained from CLIP\cite{radford2021learning}, then presents the proposed Weight Compensation Block (WCB), Noise-pair Filter Block (NFB), and the Soft-label based Noise Contrastive Estimation (NCE) loss in Section \ref{Weight Compensation}, \ref{fortune}, and \ref{nce}, respectively.

\subsection{CLIP Token Embeddings}\label{clip}
The VLP model, CLIP \cite{radford2021learning}, is known for its ability to cope with downstream cross-modal tasks. Our modification text input and the image inputs (Reference 
image, Target image) are encoded using the CLIP’s encoders $\mathbf{\Phi}_{txt}$ and $\mathbf{\Phi}_{img}$ in Fig. \ref{workflow2}, respectively, where the modification text is segmented into n+2 word-level embeddings and the image data is partitioned into m+1 patch-level embeddings, denoted as $\mathcal{T}^{mod}:\{\mathcal{T}^{mod}_{sot}, \mathcal{T}^{mod}_{1}, \mathcal{T}^{mod}_{2} ... \mathcal{T}^{mod}_{eot}\} \in \mathcal{R}^{(n+2) \cdot d}$, where $\mathcal{T}^{mod}_{sot}, \mathcal{T}^{mod}_{eot}$ are the start of token and end of token; $\mathcal{I}^{ref}:\{\mathcal{I}^{ref}_{cls}, \mathcal{I}^{ref}_{1}, \mathcal{I}^{ref}_{2} ... \mathcal{I}^{ref}_{m}\} \in \mathcal{R}^{(m+1) \cdot d}$; $\mathcal{I}^{tar}:\{\mathcal{I}^{tar}_{cls}, \mathcal{I}^{tar}_{1}, \mathcal{I}^{tar}_{2} ... \mathcal{I}^{tar}_{m}\} \in \mathcal{R}^{(m+1) \cdot d}$, where $\mathcal{I}^{ref}_{cls}$, $\mathcal{I}^{tar}_{cls}$ are the global token embeddings. Utilizing CLIP as our backbone, the cross-modal encoding provides valid feature understanding for the subsequent blocks.

\subsection{Weight Compensation Block}\label{Weight Compensation}
 Previous methods \cite{radford2021learning} often utilize the global token embeddings to represent the whole image or text features, regardless of the word-level and patch-level attention maps for subsequent processing. Yet, not all words gain equal significance, and certain image regions at the patch level may lack meaningful semantic content. To address this, we present the Weight Compensation Block (WCB), which enhances the model's capacity to capture essential semantic information while mitigating the impact of irrelevant patches and words.

Initially, the weight relocation process can be calculated as:
\begin{equation}   
\begin{aligned}
\mathcal{T}^{mod*}=\mathcal{T}^{mod} \otimes \text{Att}^{mod}&,\mathcal{I}^{ref*}=\mathcal{I}^{ref}\otimes\text{Att}^{ref},\\ \mathcal{I}^{tar*}&=\mathcal{I}^{tar}\otimes\text{Att}^{tar}, 
\end{aligned}
\end{equation}
where $\text{Att}^{mod}$, $\text{Att}^{ref}$, and $\text{Att}^{tar}$ denoted as the attention maps of the modification text, reference image, and target image.
After assigning the diverse patches and words with diverse weights, we fuse them with the global token
\begin{equation}
    \begin{aligned}
        \mathcal{T}^{mod}_{wcb} = \text{MaxPooling}(\text{MLP}(\mathcal{T}^{mod*})) \oplus \mathcal{T}^{mod}_{eot},\\
        \mathcal{I}^{ref}_{wcb} = \text{MaxPooling}(\text{MLP}(\mathcal{I}^{ref*})) \oplus \mathcal{I}^{ref}_{cls},\\
        \mathcal{I}^{tar}_{wcb} = \text{MaxPooling}(\text{MLP}(\mathcal{I}^{tar*})) \oplus \mathcal{I}^{tar}_{cls}
    \end{aligned}
\end{equation}
to get the overall weight compensation embeddings, the MLP refers to the Multi-Layer Perceptron.

Generally, the Weight Compensation Block(WCB) adjusts the weights assigned to output embeddings from CLIP, reducing the influence of irrelevant words and patches, and leading to more accurate and contextually relevant processing.

\subsection{Noise-pair Filter Block}\label{fortune}
To alleviate the detrimental effects of partially matched and mismatched pairs, it is essential to filter these potential discrepancies from the training data to prevent the introduction of false supervisory signals. Typically, matched pairs yield lower loss, whereas the noise pairs tend to result in higher loss. Driven by this and following \cite{qin2024noisy}, we propose the Noise-pair Filter Block (NFB). Its core concept employs the Gaussian Mixture Model (GMM) to model the per-batch loss distributions derived from the NCE loss function in the noise alignment module in Fig. \ref{workflow2}, aiming to further pinpoint the mismatched and partially matched pairs.  

To prepare two of the four inputs of NFB, we first have
\begin{equation}
    \begin{aligned}
        \mathcal{Q} = \text{MLP}(\mathcal{T}^{mod}_{eot},\mathcal{I}^{ref}_{cls}),
        \mathcal{Q}_{wcb} = \text{MLP}(\mathcal{T}^{mod}_{wcb},\mathcal{I}^{ref}_{wcb}),
    \end{aligned}
\end{equation}
where $\mathcal{Q}$ and $\mathcal{Q}_{wcb}$ are the integration of multi-modal query embeddings from the outputs of global embeddings outputs of CLIP encoders and the outputs of the Weight Compensation Blocks respectively. For the rest of the two inputs are $\mathcal{I}^{tar}_{cls}$ and $\mathcal{I}^{tar}_{wcb}$. The four inputs then formulate as two types of pairs input into the noise alignment module and defining the per-batch loss as:
\begin{equation}
    \begin{aligned}
        \ell = \mathcal{L}_{NCE}(\mathcal{Q}, \mathcal{I}^{tar}),
        \ell_{wcb} = \mathcal{L}_{NCE}(\mathcal{Q}_{wcb}, \mathcal{I}^{tar}_{wcb}).
    \end{aligned}
\end{equation}
Then, the $\ell$ and $\ell_{wcb}$ will be fed into the GMM to distinguish the matched pairs and the noise pairs. The Expectation-Maximization algorithms \cite{huang2021learning} 
\begin{equation}
    \begin{aligned}        
        p\left(k\mid\ell^i\right)&=p(k) p\left(\ell^i\mid k\right) / p\left(\ell^i\right),\\
        p\left(k\mid\ell^i_{wcb}\right)&=p(k) p\left(\ell^i_{wcb}\mid k\right) / p\left(\ell^i_{wcb}\right)
    \end{aligned}
\end{equation}
are used to optimize the GMM and compute the posterior probability, where $k \in \{0, 1\}$ is used to indicate the $i$-th pair to be matched or mismatched. We first generate the matched pair set and the mismatched pair set using $p\left(k=0 \mid \ell^i\right)$ and $p\left(k=0 \mid \ell^i_{wcb}\right)$ and $\theta$ = 0.5, articulated as:
\begin{equation}
    \begin{aligned}
        \mathcal{S}^{match} = \{(\mathcal{Q}, \mathcal{I}^{tar})\mid p\left(k=0 \mid \ell^i \right) > \theta\}, \\
        \mathcal{S}^{mis} = \{(\mathcal{Q}, \mathcal{I}^{tar})\mid p\left(k=0 \mid \ell^i \right) \leq \theta\}, \\
        \mathcal{S}^{match}_{wcb} = \{(\mathcal{Q}_{wcb}, \mathcal{I}^{tar}_{wcb})\mid p\left(k=0 \mid \ell^i \right) > \theta\}, \\
        \mathcal{S}^{mis}_{wcb} = \{(\mathcal{Q}_{wcb}, \mathcal{I}^{tar}_{wcb})\mid p\left(k=0 \mid \ell^i \right) \leq \theta\}. \\
    \end{aligned}
\end{equation}
The matched pairs $\mathcal{S}^{m} \leftarrow \mathcal{S}^{match} \cup \mathcal{S}^{match}_{wcb}$, and the mismatched pairs $\mathcal{S}^{u} \leftarrow \mathcal{S}^{mis} \cap \mathcal{S}^{mis}_{wcb}$. For the partially matched pair set, it can be derived from $\mathcal{S}^{m}, \mathcal{S}^{u}$, formulating as:
\begin{equation}
    \begin{aligned}
\mathcal{S}^{p} = \mathcal{S}^{mis} \cup \mathcal{S}^{mis}_{wcb} - \mathcal{S}^{mis} \cap \mathcal{S}^{mis}_{wcb}.
    \end{aligned}
\end{equation}

So far the noise pairs and the matched pairs are assigned to $\mathcal{S}^{u}, \mathcal{S}^{p}, \text{and}\ \mathcal{S}^{m}$ the sample sets correspondingly.

\subsection{Soft-label based NCE loss}\label{nce}
Based on the sample sets from NFB, we generate each pair with a soft-label
\begin{equation}
    l^{soft}= \begin{cases}1, & \text { if } \text{Sample} \in \mathcal{S}^{m}; \\ 0, & \text { if } \text{Sample} \in \mathcal{S}^{u}\ \text{or}\ \mathcal{S}^{p}, \end{cases}
\end{equation}
to avoid involving the noise pairs within training, beneficial for the model. Lastly, we design a simple function to train our model, which only uses the matched pairs in $\mathcal{S}^{m}$ for training.
\begin{equation}
\small
\begin{aligned}
    \mathcal{L}^{soft}_{NCE}=\frac{1}{B}\sum_{i=1}^Bl^{soft}(-\log \left\{\frac{\exp \left\{\cos \left(\mathcal{Q}_i, \mathcal{I}^{tar}_{j}\right) / \tau\right\}}{\sum_{j=1}^B \exp \left\{\cos \left(\mathcal{Q}_i, \mathcal{I}^{tar}_{j}\right) / \tau\right\}}\right\}) \\
    + \frac{1}{B} \sum_{i=1}^Bl^{soft}(-\log \left\{\frac{\exp \left\{\cos \left(\mathcal{Q}_{cali}, \mathcal{I}^{tar}_{calj}\right) / \tau\right\}}{\sum_{j=1}^B \exp \left\{\cos \left(\mathcal{Q}_{cali}, \mathcal{I}^{tar}_{calj}\right) / \tau\right\}}\right\}),
\end{aligned}
\end{equation}
where $\mathcal{L}^{soft}_{NCE}$ is the overall loss for our training process.
\begin{table*}[!t]
\caption{Model performance comparison on the Fashion-IQ and Shoes datasets.}\label{tab1}
\resizebox{1\textwidth}{!}{
\begin{tabular}{cccccccccc|cccc}
\midrule
\multicolumn{1}{c|}{\multirow{3}{*}{Method}} & \multicolumn{9}{c|}{Fashion-IQ}                                                                                                                      & \multicolumn{4}{c}{Shoes}                                                                                         \\ \cmidrule{2-14} 
\multicolumn{1}{c|}{}                        & \multicolumn{2}{c}{Dresses} & \multicolumn{2}{c}{Shirts} & \multicolumn{2}{c}{Top\&Tees} & \multicolumn{2}{c|}{Average}      & \multirow{2}{*}{Avg.} & \multirow{2}{*}{R@1} & \multirow{2}{*}{R@10} & \multicolumn{1}{c|}{\multirow{2}{*}{R@50}} & \multirow{2}{*}{Avg.} \\ \cmidrule{2-9}
\multicolumn{1}{c|}{}                        & R@10         & R@50         & R@10         & R@50        & R@10          & R@50          & R@10  & \multicolumn{1}{c|}{R@50} &                       &                      &                       & \multicolumn{1}{c|}{}                      &                       \\ \midrule
TIRG\cite{vo2019composing}                                         & 14.13        & 34.61        & 13.10        & 30.91       & 14.97         & 34.37         & 14.01 & 33.30                     & 23.66                 & 12.60                & 45.45                 & 69.39                                      & 42.48                 \\
VAL\cite{chen2020image}                                          & 22.38        & 44.15        & 22.53        & 44.00       & 27.53         & 51.68         & 24.15 & 46.61                     & 35.38                 & 16.49                & 49.12                 & 73.53                                      & 46.38                 \\
ARTEMIS\cite{delmas2022artemis}                                      & 25.68        & 51.05        & 21.57        & 44.13       & 28.59         & 55.06         & 25.28 & 50.08                     & 37.68                 & 18.72                & 53.11                 & 79.31                                      & 50.38                 \\
EER\cite{zhang2022composed}                                          & 30.02        & 55.44        & 25.32        & 49.87       & 33.20         & 60.34         & 29.51 & 55.22                     & 42.37                 & 20.05                & 56.02                 & 79.94                                      & 52.00                 \\
CRR\cite{zhang2022comprehensive}                                          & 30.41        & 57.11        & 30.73        & 58.02       & 33.67         & 64.48         & 31.60 & 59.87                     & 45.74                 & 18.41                & 56.38                 & 79.92                                      & 51.57                 \\
CRN\cite{yang2023composed}                                          & 32.67        & 59.30        & 30.27        & 56.97       & 37.74         & 65.94         & 33.56 & 60.74                     & 47.15                 & 18.92                & 54.55                 & 80.04                                      & 51.17                 \\
CMAP\cite{golge2014conceptmap}                                         & 36.44        & 64.25        & 34.83        & 60.06       & 41.79         & 69.12         & 37.64 & 64.42                     & 51.03                 & 21.48                & 56.18                 & 81.14                                      & 52.93                 \\
CLIP4CIR\cite{baldrati2022effective}                                     & 31.63        & 56.67        & 36.36        & 58.00       & 38.19         & 62.42         & 35.39 & 59.03                     & 47.21                 & 21.42                & 56.69                 & 81.52                                      & 53.21                 \\
FAME-ViL\cite{han2023fame}                                     & 42.19        & 67.38        & 47.64        & 68.79       & 50.69         & 73.07         & 46.84 & 69.75                     & 58.30                 &-                      &-                       &-                                            &-                       \\
Prog.Lrn.\cite{zhao2022progressive}                                     & 38.18        & 64.50        & 48.63        & 71.54       & 52.32         & 76.90         & 46.37 & 70.98                     & 58.68                 & 22.88                & 58.83                 & 84.16                                      & 55.29                 \\
SPIRIT\cite{chen2024spirit}                                       & 39.86        & 64.30        & 44.11        & 65.60       & 47.68         & 71.70         & 43.88 & 67.20                     & 55.54                 & -                    & 56.90                 & 81.49                                      & -                     \\
BKIP4CIR\cite{liu2024bi}                                     & 42.09        & 67.33        & 41.76        & 64.28       & 46.61         & 70.32         & 43.49 & 67.31                     & 55.40                 &-                      &-                       &-                                            &-                       \\
DQU-CIR\cite{dqu_cir}                                      & \underline{51.51}        & \underline{73.99}        & \underline{53.34}        & \textbf{73.11}       & \underline{58.21}         & \textbf{79.07}         & \underline{54.35} & \underline{75.39}                     & \underline{64.87}                 & \underline{31.33}                & \underline{68.41}                 & \underline{88.50}                                      & \underline{62.76}                 \\ \midrule
NCL-CIR                                   & \textbf{51.71 }       & \textbf{74.46}        & \textbf{53.43}        & \underline{72.95}       & \textbf{58.45}       & \underline{78.86}         & \textbf{54.53} & \textbf{75.42}                     & \textbf{64.98}                 & \textbf{33.10}               & \textbf{69.41}                 & \textbf{88.87}                                      & \textbf{63.80}                 \\ \midrule
\end{tabular}
}
\raggedright
\textbf{*}{The best result for each model on each evaluation metric is highlighted in bold and the second-best result is underlined (\% is omitted).}
\end{table*}
\section{Experimental Result}
\subsection{Experiments settings}
\textbf{Two types of experiments:} (1) Comparative study, and (2) Ablation study, are designed to comprehensively evaluate the efficacy of NCL-CIR. \textbf{Two benchmark datasets for CIR:} (1) Fashion-IQ\cite{wu2021fashion}, composed of three categories (Dresses, Shirts, Top\&Tees), with 18,000 triplets for training and 6,016 for testing, (2) Shoes \cite{guo2018dialog}, utilizing 10,000 images for training and 4,658 for testing. We employ the pre-processed method \cite{dqu_cir} without data augment before it is input to the CLIP encoder. \textbf{Detailed settings: } Experiments are conducted using an NVIDIA A100-80G graphics card, under the batch size of 16, with the learning rate of WCB set to be $1 \times 10^{-3}$, while the other components are set to be $1 \times 10^{-6}$ initially. \textbf{Evaluation metrics:} In accordance with the previous work, we follow standard evaluation procedures for each dataset to ensure a fair comparison, with the Fashin-IQ and Shoes employ \{Recall@10 (R@10), R@50\}, \{R@1, R@10, R@50\}.

\subsection{Comparative study}
Tab. \ref{tab1} presents the comparative results on the Fashion-IQ dataset and Shoes dataset, respectively. On the Fashion-IQ dataset, our method outperforms existing methods on most metrics, indicating its effectiveness. Additionally, it confirms that there are indeed some mismatched or partially matched sample pairs (noise pairs) in the dataset. We primarily compared several state-of-the-art methods. Existing models mainly fall into two categories: those based on traditional approaches (e.g. \cite{vo2019composing,chen2020image,delmas2022artemis}) and those based on VLP methods (e.g.\cite{baldrati2022effective, liu2024bi, dqu_cir, chen2024spirit,han2022fashionvil}). Compared to pre-trained models that are trained on large datasets, traditional models tend to be somewhat less accurate. This also demonstrates that large VLP models play an indispensable role in downstream tasks. By comparing the result of NCL-CIR and DQU-CIR with the rest of the methods, these two methods yield a marginal effect (10\%) due to the utilization of the data augmentation method \cite{dqu_cir}. By comparing NCL-CIR with DQU-CIR, we further elaborate the augmented data, by having weight compensation in SCB to make full use of the attention maps from the encoder providing compensatory embeddings and NFB taking the noise pairs and matched pairs into consideration, eventually having an improved of 0.13\%, 1.04\% in the Fashion-IQ and Shoes datasets averagely on the used evaluation metrics.


\subsection{Ablation study}
\begin{table}[!t]
\caption{Ablation study of NCL-CIR's key components - WCB and NFB, on the Shoes dataset.}\label{tab3}
\resizebox{0.5\textwidth}{!}{
\begin{tabular}{c|cccc}
\midrule
\multirow{2}{*}{Method} & \multirow{2}{*}{R@1} & \multirow{2}{*}{R@10} & \multirow{2}{*}{R@50} & \multirow{2}{*}{Avg.} \\
 &  &  &  &  \\ \midrule
Baseline w/o (WCB, NFB) &31.49  &68.20  &87.40  &62.64  \\
Baseline+WCB w/o NFB &32.18  &68.51  &88.21  &62.91  \\
Baseline+NFB w/o WCB &32.52 &68.99 &88.36 &63.33 \\ \midrule
NCL-CIR &\textbf{33.10} &\textbf{69.41} &\textbf{88.87} &\textbf{63.80} \\ \midrule
\end{tabular}
}
\end{table}
\vspace{0.1cm}
The ablation studies encompass two key components to assess the effectiveness of the NCL-CIR approach. As illustrated in \ref{tab3}, when compared to the baseline model, the incorporation of either the WCB or NFB enhances model performance, particularly with NFB, which shows a 1.03\% increase in R@1, demonstrating its efficacy in filtering out noise pairs while preserving matched pairs. Furthermore, with the synergistic combination of WCB and NFB, NCL-CIR achieves an R@1 score of 33.10\%, utilizing the original input pairs from the encoder alongside the refined output pairs from WCB. This setup facilitates more accurate feature representations, enabling more effective filtering and preserving in NFB.

\section{Conclusion}
This paper introduces the Noise-aware Contrastive Learning for Composed Image Retrieval (NCL-CIR). The network features an integrated design, with the Weight Compensation Block (WCB) addressing local information neglect from the CLIP encoder. The Noise-pair Filter Block (NFB) processes multi-scale pair-wise embeddings, ensuring effective filtering of the noise pairs (partially matched and mismatched pairs) while preserving the matched pairs. Additionally, the soft labels generated by the NFB allow for a more nuanced training approach for distinct matched pairs within the loss function. The effectiveness of NCL-CIR is validated through comparative results and ablation studies.

\newpage
\bibliographystyle{IEEEtran}
\bibliography{icassp}
\vspace{12pt}
\end{document}